\documentclass[11pt,a4paper]{article}
\usepackage[hyperref]{emnlp2020}
\usepackage{times}
\usepackage{latexsym}

\usepackage{microtype}

\aclfinalcopy % Uncomment this line for the final submission

\usepackage{times}
\usepackage{latexsym}

\usepackage{url}

\usepackage{amsmath,amsfonts,bm}

\def\eqref#1{equation~\ref{#1}}
\def\1{\bm{1}}

\DeclareMathAlphabet{\mathsfit}{\encodingdefault}{\sfdefault}{m}{sl}
\SetMathAlphabet{\mathsfit}{bold}{\encodingdefault}{\sfdefault}{bx}{n}

\newcommand{\softmax}{\mathrm{softmax}}

\usepackage{url}

\usepackage{amssymb}
\usepackage{wrapfig,lipsum,booktabs}
\usepackage{amsmath}
\usepackage{cleveref}
\usepackage{booktabs}
\usepackage{bm}
\usepackage{algorithm}
\usepackage{algpseudocode}
\usepackage{graphicx}
\usepackage{nccmath}
\usepackage{longtable}
\usepackage{caption}
\usepackage{makecell}
\usepackage{subcaption}
\usepackage{tabularx}
\usepackage{xcolor}
\newcommand{\B}[1] {\boldsymbol{#1}}

\def\bm{{\B{m}}}

\definecolor{darkpastelgreen}{rgb}{0.01, 0.75, 0.24}
\definecolor{seagreen}{rgb}{0.18, 0.55, 0.34}
\definecolor{mountainmeadow}{rgb}{0.19, 0.73, 0.56}
\definecolor{forestgreen}{rgb}{0.13, 0.55, 0.13}
\definecolor{fireenginered}{rgb}{0.81, 0.09, 0.13}
\definecolor{ballblue}{rgb}{0.13, 0.67, 0.8}
\definecolor{bleudefrance}{rgb}{0.19, 0.55, 0.91}

\makeatletter
\newcommand{\thickhline}{%
    \noalign {\ifnum 0=`}\fi \hrule height 1pt
    \futurelet \reserved@a \@xhline
}

\newcommand{\thickerhline}{%
    \noalign {\ifnum 0=`}\fi \hrule height 2pt
    \futurelet \reserved@a \@xhline
}

\makeatother

\title{A Hierarchical Network for Abstractive Meeting Summarization with Cross-Domain Pretraining}

\author{Chenguang Zhu\thanks{$\;\;$Equal contribution}$\;$, Ruochen Xu$^*$, Michael Zeng, Xuedong Huang \\
Microsoft Cognitive Services Research Group\\
  \texttt{\{chezhu,ruox,nzeng,xdh\}@microsoft.com} \\}
\date{}

\begin{document}
\maketitle

\begin{abstract}
With the abundance of automatic meeting transcripts, meeting summarization is of great interest to both participants and other parties. Traditional methods of summarizing meetings depend on complex multi-step pipelines that make joint optimization intractable. Meanwhile, there are a handful of deep neural models for text summarization and dialogue systems. However, the semantic structure and styles of meeting transcripts are quite different from articles and conversations. In this paper, we propose a novel abstractive summary network that adapts to the meeting scenario. We design a hierarchical structure to accommodate long meeting transcripts and a role vector to depict the difference among speakers. Furthermore, due to the inadequacy of meeting summary data, we pretrain the model on large-scale news summary data.
Empirical results show that our model outperforms previous approaches in both automatic metrics and human evaluation. For example, on ICSI dataset, the ROUGE-1 score increases from 34.66\% to 46.28\%.%\footnote{Code will be released on publication.}

\end{abstract}

\section{Introduction}
Meetings are a very common forum where people exchange ideas, make plans, and share information. With the ubiquity of automatic speech recognition systems come vast amounts of meeting transcripts. Therefore, the need to succinctly summarize the content of a meeting naturally arises.

\begin{table}[ht]
\small
\begin{tabular}{l}
\toprule
\textbf{Meeting Transcript (163 turns)} \\ 
\midrule
\makecell[l]{
...\\
PM: ... another point is we have to skip the \textbf{\textcolor{blue}{teletext}}, because\\ in the world of upcoming internet we think \textbf{\textcolor{blue}{teletext}} is going\\ to be a thing of the past.\\
ID: ... first about how it works. It?s really simple. Everybody\\ knows how a \textbf{\textcolor{forestgreen}{remote}} works. The user presses a button. The \\\textbf{\textcolor{forestgreen}{remote}} determines what button it is,\\
PM: ... Few buttons, we talked about that. \textbf{\textcolor{bleudefrance}{Docking station}},\\ \textbf{\textcolor{fireenginered}{LCD}}. general functions And default materials...\\
...}\\
\midrule
\textbf{Summary from our model (23 sentences)} \\
\midrule
\makecell[l]{... \\
The Project Manager announced that the project would\\ not include a
\textcolor{blue}{\textbf{teletext}} feature.\\
The Industrial Designer gave a presentation of the functions\\ of the \textbf{\textcolor{forestgreen}{remote}}.\\
The group decided on features to include in the remote, to\\include an \textbf{\textcolor{fireenginered}{LCD}} screen, and a \textbf{\textcolor{bleudefrance}{docking station}} to change\\ the layout of the interface.\\
...}\\
\bottomrule
\end{tabular}
\caption{Example excerpt of a meeting transcript and the summary generated by our model in AMI dataset. Keywords are highlighted in colors. %ME (marketing expert), 
PM (program manager) and ID (industrial designer) are roles of the speakers. The meeting transcript contains word errors and grammatical glitches as it is the result from the automatic speech recognition system.}
\label{tab:small-example-table}
\end{table}

Several methods of generating summaries for meetings have been proposed \cite{mehdad2013,murray2010,wang2013,oya2014,unsup18,acl2019_meeting}. As \citet{murray2010} points out, users prefer abstractive meeting summaries to extractive summaries. While these methods are mostly abstractive, they require complicated multi-stage machine learning pipelines, such as template generation, sentence clustering, multi-sentence compression, candidate sentence generation and ranking. As these approaches are not end-to-end optimisable, it is hard to jointly improve various parts in the pipeline to enhance the overall performance. Moreover, some components, e.g., template generation, require extensive human involvement, rendering the solution not scalable or transferrable. 

Meanwhile, many end-to-end systems have been successfully employed to tackle document summarization, such as the pointer-generator network \cite{pgnet}, reinforced summarization network \cite{rlsummarize} and memory network \cite{memnet}. These deep learning methods can effectively generate abstractive document summaries by directly optimizing pre-defined goals.

However, the meeting summarization task inherently bears a number of challenges that make it more difficult for end-to-end training than document summarization. We show an example of a meeting transcript from the AMI dataset and the summary generated by our model in Table~\ref{tab:small-example-table}.

%For instance, in AMI meeting corpus \cite{ami}, there are 4 participants per meeting, including a program manager (PM), a marketing expert (ME), a user interface designer (UI) and an industrial designer (ID). %In ICSI corpus \cite{icsi}, there are on average 6.2 participants per meeting.

First, the transcript and summary of a single meeting are usually much longer than those of a document. For instance, in CNN/Daily Mail dataset \cite{cnn}, there are on average 781 tokens per article and 56 tokens per summary, while AMI meeting corpus contains meetings with 4,757 tokens per transcript and 322 tokens per summary on average. %The example transcript in Table~\ref{tab:small-example-table} has as many as 162 turns. 
And the structure of a meeting transcript is very distinct from news articles. These challenges all prevent existing news summarization models to be successfully applied to meetings.

Second, a meeting is carried out between multiple participants. The different semantic styles, standpoints, and roles of each participant all contribute to the heterogeneous nature of the meeting transcript.

%Second, a meeting transcript is usually output from an automatic speech recognition (ASR) pipeline. Inevitably there are word errors and grammatical glitches in recognizing the colloquial language, which injects noise into the input of the meeting summary system. 

Third, compared with news, there is very limited labelled training data for meeting summary (137 meetings in AMI v.s. 312K articles in CNN/DM). This is due to the privacy of meetings and the relatively high cost of writing summaries for long transcripts.

%Third, as a meeting usually progresses according to an agenda. \citet{pgnet} shows that simply taking the first 3 sentences from a CNN news article can achieve higher ROUGE scores than several sophisticated deep learning summarization methods. This is definitely not the case for meeting transcripts.

To tackle these challenges, we propose an end-to-end deep learning framework, \textbf{H}ierarchical \textbf{M}eeting summarization \textbf{Net}work (HMNet). HMNet leverages the encoder-decoder transformer architecture \cite{transformer} to produce abstractive summaries based on meeting transcripts. To adapt the structure to meeting summarization, we propose two major design improvements.

First, as meeting transcripts are usually lengthy, a direct application of the canonical transformer structure may not be feasible. For instance, conducting the multi-head self-attention mechanism on a transcript with thousands of tokens is very time consuming and may cause memory overflow problem. Therefore, we leverage a hierarchical structure to reduce the burden of computing. As a meeting consists of utterances from different participants, it forms a natural multi-turn hierarchy. Thus, the hierarchical structure carries out both token-level understanding within each turn and turn-level understanding across the whole meeting. During summary generation, HMNet applies attention to both levels of understanding to ensure that each part of the summary stems from different portions of the transcript with varying granularities.

Second, to accommodate multi-speaker scenario, HMNet incorporates the role of each speaker\footnote{Both datasets in experiments only provide role information for each participant. In real applications, we can use a vector to represent each participant when a personal identifier is available.} to encode different semantic styles and standpoints among participants. For example, a program manager usually emphasizes the progress of the project while a user interface designer tends to focus on user experience. In HMNet, we train a role vector for each meeting participant to represent the speaker's information during encoding. This role vector is appended to the turn-level representation for later decoding.

To tackle the problem of insufficient training data for meeting summarization, we leverage the idea of pretraining \citep{bert}. We collect summarization data from the news domain and convert them into the meeting format: a group of several news articles forms a multi-person meeting and each sentence becomes a turn. The turns are reshuffled to simulate a mixed order of speakers. We pretrain the HMNet model on the news task before finetuning it on meeting summarization. Empirical results show that this cross-domain pretraining can effectively enhance the model quality.

%The training process for HMNet is end-to-end, optimizing the cross entropy of the generated summary. Therefore, HMNet makes it very convenient to jointly fine-tune each component to enhance summarization performance. To our knowledge, this is the first end-to-end deep learning method for meeting summarization entirely based on transcripts.

To evaluate our model, we employ the widely used AMI and ICSI meeting corpus \cite{ami,icsi}. Results show that HMNet significantly outperforms previous meeting summarization methods. For example, on ICSI dataset, HMNet achieves 11.62 higher ROUGE-1 points, 2.60 higher ROUGE-2 points, and 6.66 higher ROUGE-SU4 points compared with the previous best result. Human evaluations further show that HMNet generates much better summaries than baseline methods.  We then conduct ablation studies to verify the effectiveness of different components in our model.

\section{Problem Formulation}
We formalize the problem of meeting summarization as follows. The input consists of meeting transcripts $\mathcal{X}$ and meeting participants $\mathcal{P}$. Suppose there are $s$ meetings in total. The transcripts are $\mathcal{X}=\{X_1, ...,X_s\}$. Each meeting transcript consists of multiple turns, where each turn is the utterance of a participant. Thus, $X_i=\{(p_1, u_1), (p_2, u_2), ..., (p_{L_i}, u_{L_i})\}$, where $p_j\in \mathcal{P}, 1\leq j \leq L_i,$ is a participant and $u_j=(w_1,...,w_{l_j})$ is the tokenized utterance from $p_j$. The human-labelled summary for meeting $X_i$, denoted by $Y_i$, is also a sequence of tokens. For simplicity, we will drop the meeting index subscript. So the goal of the system is to generate meeting summary $Y=(y_1,...,y_n)$ given the transcripts $X=\{(p_1, u_1), (p_2, u_2), ..., (p_m, u_m)\}$.

\section{Model}
Our hierarchical meeting summarization network (HMNet) is based on the encoder-decoder transformer structure \cite{transformer}, and its goal is to maximize the conditional probability of meeting summary $Y$ given transcript $X$ and network parameters $\theta$: $P(Y|X;\theta)$.

\subsection{Encoder}
\subsubsection{Role Vector} Meeting transcripts are recorded from various participants, who may have different semantic styles and viewpoints. Therefore, the model has to take the speaker's information into account while generating summaries.

To incorporate the participants' information, we integrate the \textit{speaker role} component. In the experiments, each meeting participant has a distinct role, e.g., program manager, industrial designer. For each role, we train a vector to represent it as a fixed-length vector $r_p, 1\leq p \leq P$, where $P$ is the number of roles. Such distributed representation for a role/person has been proved to be useful for sentiment analysis \cite{chen2016learning}. This vector is appended to the embedding of the speaker's turn (Section~\ref{sec:hier}). According to the results in Section~\ref{sec:results}, the vectorized representation of speaker roles plays an important part in boosting the performance of summarization.

This idea can be extended if richer data is available in practice:
\begin{itemize}
    \item If an organization chart of participants is available, we can add in representations of the relationship between participants, e.g., manager and developers, into the network.
    \item If there is a pool of registered participants, each participant can have a personal vector which acts as a user portrait and evolves as more data about this user is collected.
\end{itemize}

\subsubsection{Hierarchical Transformer}
\label{sec:hier}
\textbf{Transformer.} Recall that a transformer block consists of a multi-head attention layer and a feed-forward layer, both followed by layer-norm with residuals: $\mbox{LayerNorm}(x + Layer(x))$, where $Layer$ can be the attention or feed-forward layer \cite{transformer}.

%In detail, the input consists of query $\bQ\in \mathbb{R}^{n\times d}$, key $\bK\in \mathbb{R}^{m\times d}$ and value $\bV\in \mathbb{R}^{m\times d}$, the attention is based on inner product of query and key:

%\begin{equation}
%    \mbox{Attention}(\bQ, \bK, \bV) = \softmax(\frac{\bQ\bK^T}{\sqrt{d}})\bV\in \mathbb{R}^{n\times d}
%\end{equation}

%In self-attention, $\bQ=\bK=\bV$. In cross-attention, $\bQ$ and $\bK=\bV$ represent the source and target of attention, respectively. 

%Multi-head attention employs the attention $h$ times, each time projecting $\bQ, \bK, \bV$ to a space of $\frac{d}{h}$ dimensions:
%\begin{align}
%    \mbox{MultiHead}(\bQ, &\bK, \bV) = \mbox{Concat}(\bA_1, \bA_2, ...,\bA_h)\bW^O \\
%    \bA_i &= \mbox{Attention}(\bQ\bW^Q_i, \bK\bW^K_i, \bV\bW^V_i)
%\end{align}

%Where the projection matrices are  $\bW^Q_i\in \mathbb{R}^{d\times \frac{d}{h}}, \bW^K_i\in \mathbb{R}^{d\times \frac{d}{h}}, \bW^V_i\in \mathbb{R}^{d\times \frac{d}{h}}$ and $\bW^O\in \mathbb{R}^{d\times d}$.

%The feed-forward network utilizes two linear transformations:
%\begin{equation}
%    \mbox{FFN}(x) = \max(0, x\bW_1 + b_1)\bW_2+b_2
%\end{equation}

As the attention mechanism is position agnostic, we append positional encoding to input vectors:

\begin{align}
    \mbox{PE}_{(i,2j)}&=\mbox{sin}(i/10000^{\frac{2j}{d}})\\
    \mbox{PE}_{(i,2j+1)}&=\mbox{cos}(i/10000^{\frac{2j}{d}}),
\end{align}
where $\mbox{PE}_{(i,j)}$ stands for the $j$-th dimension of positional encoding for the $i$-th word in input sequence. We choose sinusoidal functions as they can extend to arbitrary input length during inference.

In summary, a transformer block on a sequence of $n$ input embeddings can generate $n$ output embeddings of the same dimension as input. Thus, multiple transformer blocks can be sequentially stacked to form a transformer network:
\begin{equation}
    \mbox{Transformer}(\{x_1, ..., x_n\}) = \{y_1, ..., y_n\}
\end{equation}

\begin{figure*}[htbp]
\centering
\vspace{2\baselineskip}
\hspace*{1.2cm}\includegraphics[scale=0.55,trim=0cm 0cm 0cm 3cm]{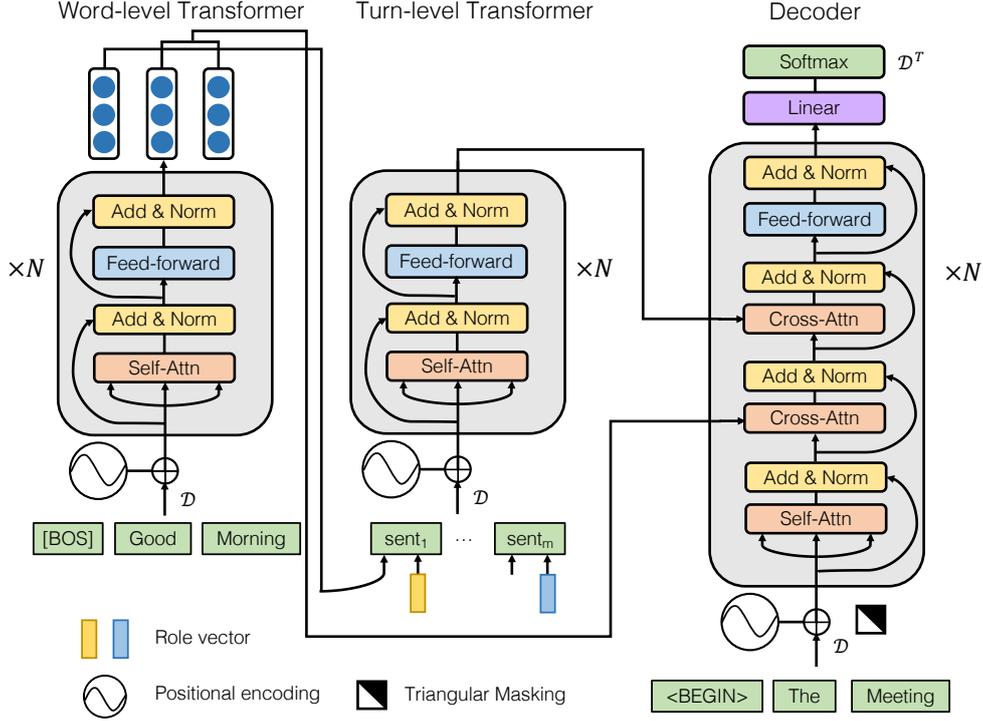}
\vspace{-2\baselineskip}
\caption{Hierarchical Meeting Summary Network (HMNet) model structure. [BOS] is the special start token inserted before each turn, and its encoding is used in turn-level transformer encoder. Other tokens' encodings enter the cross-attention module in decoder.}
\label{fig:model}
\end{figure*}

\textbf{Long transcript problem.} As the canonical transformer has the attention mechanism, its computational complexity is quadratic in the input length. Thus, it struggles to handle very long sequences, e.g. 5,000 tokens. However, meeting transcripts are usually fairly long, consisting of thousands of tokens. 

We note that meetings come with a natural multi-turn structure with a reasonable number of turns, e.g. 289 turns per meeting on average in AMI dataset. And the number of tokens in a turn is much less than that in the whole meeting. Therefore, we employ a two-level transformer structure to encode the meeting transcript.

\textbf{Word-level Transformer}. The word-level transformer processes the token sequence of one turn in the meeting. We encode each token in one turn using a trainable embedding matrix $\mathcal{D}$.
% initialized by GloVe \cite{glove}.
Thus, the $j$-th token in the $i$-th turn, $w_{i,j}$, is associated with a uniform length vector $\mathcal{D}(w_{i,j})=g_{i,j}$. To incorporate syntactic and semantic information, we also train two embedding matrices to represent the part-of-speech (POS) and entity (ENT) tags. Therefore, the token $w_{i,j}$ is represented by the vector $x_{i,j}=[g_{i,j}; POS_{i,j}; ENT_{i,j}]$. Note that we add a special token $w_{i,0}$=[BOS] before the sequence to represent the beginning of a turn. Then, we denote the output of the word-level transformer as follows: $\mbox{Word-Transformer}(\{x_{i,0}, ..., x_{i,L_i}\}) = \{x^{\mathcal{W}}_{i,0}, ..., x^{\mathcal{W}}_{i,L_i}\}$.

\textbf{Turn-level Transformer}. The turn-level transformer processes the information of all $m$ turns in a meeting. To represent the $i$-th turn, we employ the output embedding of the special token [BOS] from the word-level transformer, i.e. $x^{\mathcal{W}}_{i,0}$. Furthermore, we concatenate it with the role vector of the speaker for this turn, $p_i$. It follows that the output of the turn-level transformer is: $\mbox{Turn-Transformer}(\{[x^{\mathcal{W}}_{1,0}; p_1], ..., [x^{\mathcal{W}}_{m,0}; p_m]\}) = \{x^{\mathcal{T}}_1, ..., x^{\mathcal{T}}_m\}$. 

\subsection{Decoder}
The decoder is a transformer to generate the summary tokens. The input to the decoder transformer contains the $k-1$ previously generated summary tokens $\hat{y}_1, ..., \hat{y}_{k-1}$. Each token is represented by a vector using the same embedding matrix $\mathcal{D}$ as the encoder, $\mathcal{D}(\hat{y}_i)=g_i$.

The decoder transformer uses a lower triangular mask to prevent the model to look at future tokens. Moreover, the transformer block includes two cross-attention layers. After self-attention, the embeddings first attend with token-level outputs $\{x^{\mathcal{W}}_{i,j}\}_{i=1,j=1}^{m,L_i}$, and then with turn-level outputs $\{x^{\mathcal{T}}_i\}_{i=1}^m$, each followed by layer-norm. This makes the model attend to different parts of the inputs with varying scales at each inference step.

The output of the decoder transformer is denoted as:
$\mbox{Decoder-Transformer}(\{g_1,...,g_{k-1}\})=\{v_1,...,v_{k-1}\}$.

To predict the next token $\hat{y}_{k}$, we reuse the weight of embedding matrix $\mathcal{D}$ to decode $v_{k-1}$ into a probability distribution over the vocabulary: %$\bW_{\mathcal{D}}$\footnote{Suppose $\mathcal{D}\in \mathbb{R}^{|V|\times d}$, its transposed weights $\bW_{\mathcal{D}}\in \mathbb{R}^{d\times |V|}$.} :

\begin{equation}
    P(\hat{y}_k|\hat{y}_{<k},X)=\softmax(v_{k-1}\mathcal{D}^T)
\end{equation}

We illustrate the Hierarchical Meeting summary Network (HMNet) in Fig.~\ref{fig:model}.

\textbf{Training}. During training, we seek to minimize the cross entropy:
\begin{equation}
    L(\theta) = -\frac{1}{n}\sum_{k=1}^n logP(y_k|y_{<k}, X)
\end{equation}

We use teacher-forcing in decoder training, i.e. the decoder takes ground-truth summary tokens as input.

\textbf{Inference}. During inference, we use beam search to select the best candidate. The search starts with the special token $\langle\mbox{BEGIN}\rangle$. We employ the commonly used trigram blocking \citep{rlsummarize}: during beam search, if a candidate word would create a trigram that already exists in the previously generated sequence of the beam, we forcibly set the word's probability to 0. Finally, we select the summary with the highest average log-likelihood per token.

\subsection{Pretraining}
As there is limited availability of meeting summarization data, we propose to utilize summary data from the news domain to pretrain HMNet. This can warm up model parameters on summarization tasks. However, the structure of news articles is very different from meeting transcripts. Therefore, we transform news articles into the meeting format.

We concatenate every $M$ news articles into an $M$-people meeting, and treat each sentence as a single turn. The sentences from article $i$ is considered to be utterances from the $i$-th speaker, named as [Dataset-$i$]. %To let the model learn to utilize speaker information during pretraining, we use the combination of the dataset name and the article index as the speaker's name. 
For instance, for each XSum meeting, the speakers' names are [XSum-1] to [XSum-$M$]. To simulate the real meeting scenario, we randomly shuffle all the turns in these pseudo meetings. The target summary is the concatenation of the $M$ summaries.
%This is to make the model aware of the differences in the news distribution from different datasets.

We pretrain HMNet model with a large collection of news summary data (details in Section~\ref{sec:data}), and then finetune it on real meeting summary task.

\begin{table*}[htbp]
\centering
\begin{tabular}{l|lll|lll}
\thickhline
  & \multicolumn{3}{c|}{AMI} & \multicolumn{3}{c}{ICSI} \\
  \hline
  Model & ROUGE-1 & R-2 & R-SU4 &  ROUGE-1 & R-2 & R-SU4  \\
 \hline
Random & 35.13 & 6.26 & 13.17 & 29.28 & 3.78 & 10.29 \\
%Longest Greedy & 33.35 & 5.11 & 12.15 & 30.23 & 4.27 & 10.90 \\
Template & 31.50 & 6.80& 11.40 & / & / & / \\
%CoreRank Submodular & 36.13 & 7.33 & 14.18 & 29.82 & 4.00 & 10.61 \\
%PageRank Submodular & 36.1 & 7.42 & 14.32 & 30.4 & 4.42 & 11.14 \\
TextRank & 35.25 & 6.9 & 13.62 & 29.7 & 4.09 & 10.64 \\
ClusterRank & 35.14 & 6.46 & 13.35 & 27.64 & 3.68 & 9.77 \\
UNS & 37.86 & 7.84 & 14.71 & 31.60 & 4.83 & 11.35 \\
Extractive Oracle & 39.49 & 9.65 & 13.20 & 34.66 & 8.00 & 10.49 \\
PGNet & 40.77 & 14.87 & 18.68 & 32.00 & 7.70 & 12.46 \\
Copy from Train & 43.24 & 12.15 & 14.01 & 34.65 & 5.55 & 10.65 \\
MM (TopicSeg+VFOA)$^*$ & \textbf{53.29} & 13.51 & / & / & / & / \\
MM (TopicSeg)$^*$ & 51.53 & 12.23 & / & / & / & / \\
%HMNet & 52.09($\pm$2.35) & \textbf{19.69}($\pm$2.32) & \textbf{24.11}($\pm$4.25) & \textbf{38.38}($\pm$1.80) & \textbf{9.08}($\pm$1.50) & \textbf{16.56}($\pm$2.18) \\
\hline
% HMNet-RNN (ours) & 42.1 & 14.52 & 18.94 & 36.67 & 9.84 & 14.88\\
HMNet & 53.02 & \textbf{18.57$^{**}$} & \textbf{24.85$^{**}$} & \textbf{46.28$^{**}$} & \textbf{10.60$^{**}$} & \textbf{19.12$^{**}$} \\
\thickhline
\end{tabular}
\caption{ROUGE-1, ROUGE-2, ROUGE-SU4 scores of generated summary in AMI and ICSI datasets. Numbers in bold are the overall best result. $^*$ The two baseline MM models require additional human annotations of topic segmentation and visual signals from cameras. $^{**}$ Results are statistically significant at level 0.05.} 
\label{table:mainresult}
\end{table*}

\section{Experiment}
\label{exp}
\subsection{Datasets}
\label{sec:data}
We employ the widely used AMI \cite{ami} and ICSI \cite{icsi} meeting corpora. The two datasets contain meeting transcripts from automatic speech recognition (ASR), respectively. We follow \citet{unsup18} to use the same train/development/test split: 100/17/20 for AMI and 43/10/6 for ICSI. %Table~\ref{table:datasize} shows the details of train/dev/test split.
Each meeting has an abstractive summary written by human annotators. Furthermore, each participant has an associated role, e.g. project manager, marketing expert\footnote{We select the Scenario Meetings of AMI as in \citet{unsup18}}. Since there is only one speaker per role in each meeting and no other speaker identification information, we use a single role vector to model both speaker and role information simultaneously.

In AMI, there are on average 4,757 words with 289 turns in the meeting transcript and 322 words in the summary. In ICSI, there are on average 10,189 words with 464 turns in the meeting transcript and 534 words in the summary. As the transcript is produced by the ASR system, there is a word error rate of 36\% for AMI and 37\% for ICSI \citep{unsup18}.

The pretraining is conduct on the news summarization datasets CNN\slash{DailyMail} \citep{cnn}, NYT \citep{nyt} and XSum \citep{xsum}, containing 312K, 104K and 227K article-summary pairs. We take the union of three datasets for the pretraining.
We choose groups of $M=4$ news articles to match the 4-speaker setting in AMI dataset.
%We use NLTK toolkit \citep{nltk} to tokenize each article into sentences.
These converted meetings contain on average 2,812 words with 128 turns and 176 words in the summary.
% \chezhu{more details here...}

%\begin{table}[htbp]
%\centering
%\begin{tabular}{lccc}
%\toprule
% Dataset & Training & Development & Test \\
%  \midrule
%AMI & 100 & 17 & 20 \\
%ICSI & 43 & 10 & 6 \\
%\bottomrule
%\end{tabular}
%\caption{Number of meetings in train/dev/test split for AMI and ICSI corpora.} 
%\label{table:datasize}
%\end{table}

\subsection{Baseline models}
%\textbf{Random} and \textbf{Longest Greedy} \cite{random}
For comparison, we select a variety of baseline systems from previous literatures: the basic baselines \textbf{Random} \citep{random} and \textbf{Copy from Train}, which randomly copies a summary from the training set as the prediction\footnote{To reduce variance, for each article, we randomly sample 50 times and report the averaged metrics.}; the template-based method \textbf{Template} \cite{oya2014}; the ranking systems \textbf{TextRank} \cite{textrank} and \textbf{ClusterRank} \citep{clusterrank}; the unsupervised method \textbf{UNS}; the document summarization model \textbf{PGNet}\footnote{PGNet treats the whole meeting transcript as an article and generates the summary.} \citep{pgnet}; and the multi-modal model \textbf{MM} \cite{acl2019_meeting}.

In addition, we implement %two baseline models: \textbf{HMNet-RNN}, which has the same structure of HMNet but uses LSTM of 128 hidden units instead of transformer as encoding units; 
the baseline model \textbf{Extractive Oracle}, which concatenates top sentences with the highest ROUGE-1 scores with the golden summary. The number of sentences is determined by the average length of golden summary: 18 for AMI and 23 for ICSI. %The performance of Extractive Oracle can be seen as the upper bound for extractive summarization.

%And we obtain the results of baseline models from \cite{unsup18}, which also contains detailed descriptions of these models.

\subsection{Metrics}
Following \citet{unsup18}, we employ ROUGE-1, ROUGE-2 and ROUGE-SU4 metrics \cite{rouge} to evaluate all meeting summarization models. These three metrics respectively evaluate the accuracy of unigrams, bigrams, and unigrams plus skip-bigrams with a maximum skip distance of 4. These metrics have been shown to highly correlate with the human judgment \cite{rouge}.

\subsection{Implementation Details}
% [TODO] change to pretrained settings
We employ spaCy \cite{spacy} as the word tokenizer and embed POS and NER tags into 16-dim vectors. The dimension of the role vector is 32.
%All transformers share the same hyperparameters as \textit{transformer-base} in \cite{transformer}.

All transformers have 6 layers and 8 heads in attention.
The dimension for each word is $512$ and thus the input and output dimensions of transformers $d_{model}$ are $512$ for the decoder, $512+16+16=544$ for the word-level transformer, and $512+16+16+32=576$ for the turn level transformer. For all transformers, the inner-layer always has dimensionality $d_{ff} = 4 \times d_{model}$. HMNet has 204M parameters in total. We use a dropout probability of 0.1 on all layers. 

We pretrain HMNet on news summarization data using the RAdam optimizer \cite{Liu2020On} with $\beta_1=0.9$, $\beta_2=0.999$. The initial learning rate is set to $1e-9$ and linearly increased to $0.001$ with $16000$ warmup steps. For finetuning on the meeting data, the optimization setup is the same except the initial learning rate is set to $0.0001$. We use gradient clipping with a maximum norm of 2 and gradient accumulation steps as 16.
% All transformers have 6 layers and 8 heads in attention.
% The dimension for each word is $512$ and thus the input and output dimensions of transformers $d_{model}$ are $512$ for the decoder, $512+16+16=544$ for the word-level transformer, and $512+16+16+32=576$ for the turn level transformer. For all transformers, the inner-layer always has dimensionality $d_{ff} = 4 \times d_{model}$. HMNet has 204M parameters in total. We use a dropout probability of 0.1 on all layers. 
%We present more training details in Appendix A.

\subsection{Results}
\label{sec:results}
Table~\ref{table:mainresult} shows the ROUGE scores of generated summaries in AMI and ICSI datasets. As shown, except for ROUGE-1 in AMI, HMNet outperforms all baseline models in all metrics, and the result is statistically significant at level 0.05, under paired t-test with the best baseline results. On ICSI dataset, HMNet achieves 11.62, 2.60 and 6.66 higher ROUGE points than previously best results. 

Note that MM is a multi-modal model which requires human annotation of topic segmentation (TopicSeg) and visual focus on attention (VFOA) collected from cameras, which is rarely available in practice. In comparison, our model HMNet is entirely based on transcripts from ASR pipelines. Still, on AMI dataset, HMNet outperforms MM(TopicSeg) by 1.49 points in ROUGE-1 and 6.34 points in ROUGE-2, and is higher than MM(TopicSeg+VFOA) by 5.06 points in ROUGE-2.

Moreover, HMNet significantly outperforms the document summarization model PGNet, indicating that traditional summarization models must be carefully adapted to meeting scenarios. HMNet also compares favorably to the extractive oracle, showing that human summaries are more abstractive rather than extractive for meetings. 

It's worth noting that Copy from Train obtains a surprisingly good result in both AMI and ICSI, higher than most baselines including PGNet. The reason is that the meetings in AMI and ICSI are not isolated events. Instead, they form a series of related discussions on the same project. Thus, many project keywords appear in multiple meetings and their summaries. It also explains the relatively high ROUGE scores in the evaluation. However, HMNet can focus on salient information and as a result, achieves a considerably higher score than Copy from Train baseline.

%The RNN version of HMNet, HMNet-RNN, also achieves remarkable results by outperforming many baseline models.

%In the supplementary materials, we show example summaries generated by our model and baseline models. Compared to baselines, HMNet can summarize both individual actions and group activities, similar to the reference's structure. Also, the language by HMNet is much smoother and contains fewer grammatical errors.

\textbf{Ablation Study.} Table~\ref{table:ablation} shows the ablation study of HMNet on the test set of AMI and ICSI. As shown, the pretraining on news summarization data can help increase the ROUGE-1 on AMI by 4.3 points and on ICSI by 4.0 points.  %Part-of-speech and entity embeddings are also removed, the ROUGE-1 score drops 3.8 points.
When the role vector is removed, the ROUGE-1 score drops 5.2 points on AMI and 2.3 points on ICSI. %``+Role text'' means that the role is not represented by embedding. Instead, the role name is prepended to each turn. Its ROUGE-1 is 5.6 points lower than HMNet. This shows the effectiveness of a dedicated vector representation of speaker roles.
When HMNet is without the hierarchy structure, i.e. the turn-level transformer is removed and role vectors are appended to word-level embeddings, the ROUGE-1 score drops as much as 7.9 points on AMI and 5.3 points on ICSI. Thus, all these components we propose both play an important role in the summarization capability of HMNet.

\begin{table}[tbp]
\centering
\begin{tabular}{l|ccc}
\thickhline
  Model & ROUGE-1 & R-2 & R-SU4 \\
\hline
\multicolumn{4}{c}{AMI} \\
\hline
HMNet & \textbf{53.0} & \textbf{18.6} & \textbf{24.9} \\
\quad $-$pretrain & 48.7 & 18.4 & 23.5 \\
%\quad $-$POS\&ENT & 49.3 & 18.8 & 23.5\\
\quad $-$role vector & 47.8 & 17.2 & 21.7 \\
\quad $-$hierarchy & 45.1 & 15.9 & 20.5 \\
\hline
\multicolumn{4}{c}{ICSI} \\
\hline
HMNet & \textbf{46.3} & \textbf{10.6} & \textbf{19.1} \\
\quad $-$pretrain & 42.3 & \textbf{10.6} & 17.8 \\
\quad $-$role vector & 44.0 & 9.6 & 18.2\\
\quad $-$hierarchy & 41.0 & 9.3 & 16.8 \\
\thickhline
\end{tabular}
\caption{Ablation study of HMNet.} 
\label{table:ablation}
\end{table}

\subsection{Human Evaluation}
We conduct a human evaluation of the meeting summary to assess its readability and relevance. Readability measures how fluent the summary language is, including word and grammatical error rate. Relevance measures how well the summary sums up the main ideas of the meeting. 

As MM model \cite{acl2019_meeting} does not have summarization text or trained model available, we compare the results of HMNet and UNS \cite{unsup18}. For each meeting in the test set of AMI and ICSI, we have 5 human evaluators from Amazon Mechanical Turk label summaries from HMNet and UNS. We choose labelers with high approval rating ($>$98\%) to increase the credibility of results.

Each annotator is presented with the meeting transcript and the summaries. The annotator needs to give a score from 1 to 5 (higher is better) for readability (whether the summary consists of fluent and coherent sentences and easy to understand) and likewise for relevance (whether the summary contains important information from the meeting). The annotators need to read both the meeting transcript and the summary to give evaluations. To reduce bias, for each meeting, the two versions of summaries are randomly ordered.

Table~\ref{table:human} shows that HMNet achieves much higher scores in both readability and relevance than UNS in both datasets. And the scores for HMNet are all above 4.0, indicating that it can generate both readable and highly relevant meeting summaries.

\begin{figure}[tbp]
\centering
\includegraphics[width=7cm]{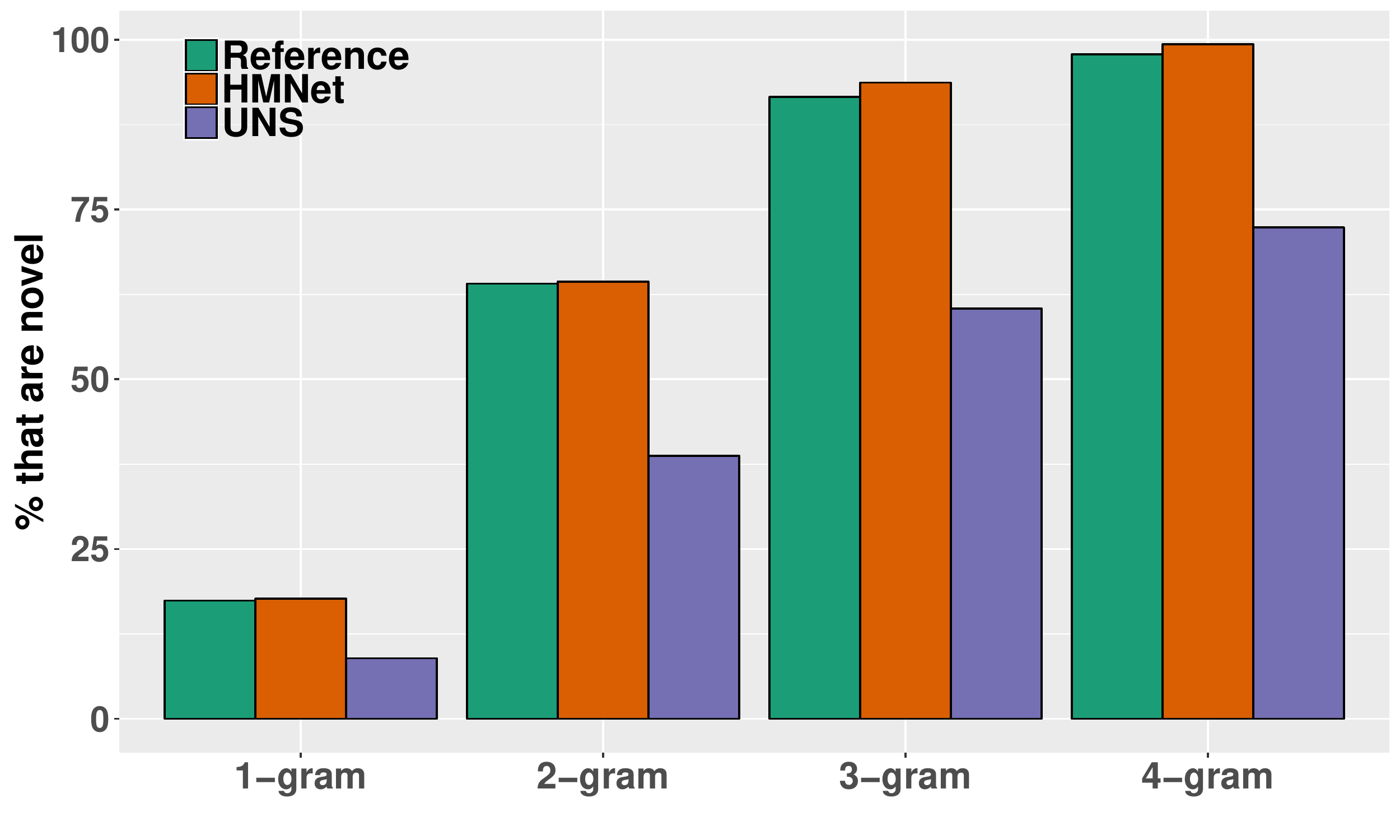}
\caption{Percentage of novel n-grams in the reference and the summaries generated by HMNet and UNS \cite{unsup18} in AMI's test set.}
\label{fig:novel}
\end{figure}

\begin{table}[tbp]
\centering
\begin{tabular}{l|c|c}
\thickhline
\textbf{Dataset} & \multicolumn{2}{c}{AMI} \\
\hline
\textbf{Source} & HMNet & UNS \\
\hline
\textbf{Readability} & \textbf{4.17} (.38) & 2.19 (.57)\\
\textbf{Relevance} & \textbf{4.08} (.45) & 2.47 (.67)\\
\hline
\textbf{Dataset} & \multicolumn{2}{c}{ICSI} \\
\hline
\textbf{Source} & HMNet & UNS \\
\hline
\textbf{Readability} & \textbf{4.24} (.20) & 2.08 (.20)\\
\textbf{Relevance} & \textbf{4.02} (.55) & 1.75 (.61)\\
\thickhline
\end{tabular}
\caption{Average scores (1-5) of readability and relevance of summaries on AMI and ICSI's test sets. Each summary is judged by 5 human evaluators. Standard deviation is shown in parenthesis.} 
\label{table:human}
\end{table}

\section{Insights}
\subsection{How abstractive is our model?}
An abstractive system can be innovative by using words that are not from the transcript in the summary. Similar to \citet{pgnet}, we measure the abstractiveness of a summary model via the ratio of novel words or phrases in the summary. A higher ratio could indicate a more abstractive system.

Fig.~\ref{fig:novel} displays the percentage of novel n-grams, i.e. that do not appear in the meeting transcript, in the summary from reference, HMNet, and UNS. As shown, both reference and HMNet summaries have a large portion of novel n-grams ($n>1$). Almost no 4-grams are copied from the transcript. In contrast, UNS has a much lower ratio of novel n-grams, because it generates a summary mainly from the original word sequence in transcripts. %It's worth noting that HMNet has a bit higher novelty ratio than the reference summary, which shows its strong abstractive ability.

%the original transcript, while the summary from UNS has many grammatic issues. The reason is that UNS generates sentences from word sequences during multi-sentence compression, which does not include any modeling of language. 

%Therefore, the end-to-end HMNet can effectively capture key points in a meeting and deliver a much smoother summary than pipeline-based UNS.

\subsection{Error Analysis}
We qualitatively examine the outputs of HMNet and summarize two major types of errors:

1. Due to the nature of long meeting transcripts, the system sometimes summarizes salient information from parts of the meeting different from the reference summaries.

2. Our system sometimes summarizes meetings at a high level (e.g. topics, decisions) and not to cover all detailed items as in the reference.

% \begin{itemize}
%     \item Due to the nature of long meeting transcripts, the system sometimes summarizes salient information from parts of the meeting different from the reference summaries.
%     % \item Our system tends to summarize meetings at a high level (e.g. topics, decisions) and not to describe details as much as that in the reference.
%     \item Our system sometimes makes repetition or grammatical errors on the list of detailed examples. This is mainly due to flexible language usage and Automatic Speech Recognition (ASR) noise in the meeting transcript.
%     %\item Some sentences from generated summaries are not grammatically correct. This is mainly due to the fact that the input transcripts are from the Automatic Speech Recognition (ASR) system and thus contain word errors, grammatical glitches, and incomplete sentences.
% \end{itemize}

\section{Related Work}
\textbf{Meeting Summarization.} There are a number of studies on generating summaries for meetings and dialogues \citep{zhao2019abstractive,liu-chen-2019-reading,chen2012integrating,liu2019topic,didi}. \citet{mehdad2013} uses utterance clustering, an entailment graph, a semantic word graph and a ranking strategy to construct meeting summaries. \citet{murray2010} and \citet{wang2013} focus on various aspects of meetings such as decisions and action items. \citet{oya2014} employs multi-sentence fusion to construct summarization templates for meetings, leading to summaries with higher readability and informativeness. Recently, \citet{unsup18} leverages a multi-sentence compression graph and budgeted submodular maximization to generate meeting summaries. %In general, these approaches first semantically group utterances, then over-generate candidate summary sentences via word graph. This is followed by ranking and selecting top sentences into the summary. 
In general, these multi-step methods make joint optimization intractable. % and any modification to one component often requires changing other components in the pipeline. 
\citet{acl2019_meeting} proposes an encoder-decoder structure for end-to-end multi-modal meeting summarization, but it depends on manual annotation of topic segmentation and visual focus, which may not be available in practice. In comparison, our model only requires meeting transcripts directly from speech recognition. 

\textbf{Document Summarization.} 
% End-to-end abstractive document summarization has received considerable attention in recent literature. 
\citet{nnlm} first introduces an attention-based seq2seq \cite{seq2seq} model to the abstractive sentence summarization task. However, the quality of the generated multi-sentence summaries for long documents is often low, and out of vocabulary (OOV) words cannot be efficiently handled. To tackle these challenges, \citet{pgnet} proposes a pointer-generator network that can both produce words from the vocabulary via a generator and copy words from the source text via a pointer. \citet{rlsummarize} further adds reinforcement learning to improve the result. \citet{bottomup} uses a content selector to over-determine phrases in source documents that helps constrain the model to likely phrases and achieves state-of-the-art results in several document summarization datasets. Recently several works on using large-scale pretrained language models for summarization are proposed and achieves very good performance  \citep{bertsum,leadbias,t5,bart,pegasus}.

%\cite{transformer_summary} utilizes a modified decoder-only transformer architecture for extractive text summarization. 

\textbf{Hierarchical Neural Architecture.} As a variety of NLP data (e.g., conversation, document) has an internal hierarchical structure, there have been many works applying hierarchical structures in NLP tasks. \citet{hier} proposes a hierarchical neural auto-encoder for paragraph and document reconstruction. It applies two levels of RNN: one on tokens within each sentence and the other on all sentences. \citet{docmodelhier} applies a hierarchical RNN language model (HRNNLM) to document modeling, which similarly encodes token-level and turn-level information for better language modeling performance. \citet{hred} puts forward a hierarchical recurrent encoder-decoder network (HRED) to model open-domain dialogue systems and generate system responses given the previous context. \citet{dailymail} proposes the hierarchical attention mechanism on word-level and turn-level in the encoder-decoder structure for abstractive document summarization. %Compared to these works, we apply a hierarchical transformer structure for meeting summarization.

\section{Conclusion}
\label{conclusion}
In this paper, we present an end-to-end hierarchical neural network, HMNet, for abstractive meeting summarization. We employ a two-level hierarchical structure to adapt to the long meeting transcript, and a role vector to represent each participant. We also alleviate the data scarcity problem by pretraining on news summarization data. Experiments show that HMNet achieves state-of-the-art performance in both automatic metrics and human evaluation. Through an ablation study, we show that the role vector, hierarchical architecture, and pretraining all contribute to the model's performance. 

For future work, we plan to utilize organizational chart, knowledge graph and topic modeling to generate better meeting summaries, which can better capture salient information from the transcript.

\section*{Acknowledgement}
We thank William Hinthorn for proof-reading this paper. We thank the anonymous reviewers for their valuable comments.

\bibliography{main}

\begin{thebibliography}{40}
\expandafter\ifx\csname natexlab\endcsname\relax\def\natexlab#1{#1}\fi

\bibitem[{Chen et~al.(2016)Chen, Xu, He, Xia, and Wang}]{chen2016learning}
Tao Chen, Ruifeng Xu, Yulan He, Yunqing Xia, and Xuan Wang. 2016.
\newblock Learning user and product distributed representations using a
  sequence model for sentiment analysis.
\newblock \emph{IEEE Computational Intelligence Magazine}, 11(3):34--44.

\bibitem[{Chen and Metze(2012)}]{chen2012integrating}
Yun-Nung Chen and Florian Metze. 2012.
\newblock Integrating intra-speaker topic modeling and temporal-based
  inter-speaker topic modeling in random walk for improved multi-party meeting
  summarization.
\newblock In \emph{Thirteenth Annual Conference of the International Speech
  Communication Association}.

\bibitem[{Devlin et~al.(2018)Devlin, Chang, Lee, and Toutanova}]{bert}
Jacob Devlin, Ming-Wei Chang, Kenton Lee, and Kristina Toutanova. 2018.
\newblock Bert: Pre-training of deep bidirectional transformers for language
  understanding.
\newblock \emph{arXiv preprint arXiv:1810.04805}.

\bibitem[{Garg et~al.(2009)Garg, Favre, Reidhammer, and
  Hakkani-T{\"u}r}]{clusterrank}
Nikhil Garg, Benoit Favre, Korbinian Reidhammer, and Dilek Hakkani-T{\"u}r.
  2009.
\newblock Clusterrank: a graph based method for meeting summarization.
\newblock \emph{Tenth Annual Conference of the International Speech
  Communication Association,}.

\bibitem[{Gehrmann et~al.(2018)Gehrmann, Deng, and Rush}]{bottomup}
Sebastian Gehrmann, Yuntian Deng, and Alexander~M Rush. 2018.
\newblock Bottom-up abstractive summarization.
\newblock In \emph{Proceedings of the 2018 Conference on Empirical Methods in
  Natural Language Processing}, pages 4098--4109.

\bibitem[{Hermann et~al.(2015)Hermann, Kocisky, Grefenstette, Espeholt, Kay,
  Suleyman, and Blunsom}]{cnn}
Karl~Moritz Hermann, Tomas Kocisky, Edward Grefenstette, Lasse Espeholt, Will
  Kay, Mustafa Suleyman, and Phil Blunsom. 2015.
\newblock Teaching machines to read and comprehend.
\newblock \emph{Advances in neural information processing systems,}, pages
  1693--1701.

\bibitem[{Honnibal and Johnson(2015)}]{spacy}
Matthew Honnibal and Mark Johnson. 2015.
\newblock An improved non-monotonic transition system for dependency parsing.
\newblock \emph{Proceedings of the 2015 Conference on Empirical Methods in
  Natural Language Processing,}, pages 1373--1378.

\bibitem[{Janin et~al.(2003)Janin, Baron, Edwards, Ellis, Gelbart, Morgan,
  Peskin, Pfau, Shriberg, Stolcke et~al.}]{icsi}
Adam Janin, Don Baron, Jane Edwards, Dan Ellis, David Gelbart, Nelson Morgan,
  Barbara Peskin, Thilo Pfau, Elizabeth Shriberg, Andreas Stolcke, et~al. 2003.
\newblock The icsi meeting corpus.
\newblock \emph{2003 IEEE International Conference on Acoustics, Speech, and
  Signal Processing, 2003. Proceedings.(ICASSP'03),}, 1:I--I.

\bibitem[{Jiang and Bansal(2018)}]{memnet}
Yichen Jiang and Mohit Bansal. 2018.
\newblock Closed-book training to improve summarization encoder memory.
\newblock In \emph{Proceedings of the 2018 Conference on Empirical Methods in
  Natural Language Processing}, pages 4067--4077.

\bibitem[{Lewis et~al.(2019)Lewis, Liu, Goyal, Ghazvininejad, Mohamed, Levy,
  Stoyanov, and Zettlemoyer}]{bart}
Mike Lewis, Yinhan Liu, Naman Goyal, Marjan Ghazvininejad, Abdelrahman Mohamed,
  Omer Levy, Ves Stoyanov, and Luke Zettlemoyer. 2019.
\newblock Bart: Denoising sequence-to-sequence pre-training for natural
  language generation, translation, and comprehension.
\newblock \emph{arXiv preprint arXiv:1910.13461}.

\bibitem[{Li et~al.(2015)Li, Luong, and Jurafsky}]{hier}
Jiwei Li, Minh-Thang Luong, and Dan Jurafsky. 2015.
\newblock A hierarchical neural autoencoder for paragraphs and documents.
\newblock \emph{arXiv preprint arXiv:1506.01057}.

\bibitem[{Li et~al.(2019)Li, Zhang, Ji, and Radke}]{acl2019_meeting}
Manling Li, Lingyu Zhang, Heng Ji, and Richard~J Radke. 2019.
\newblock Keep meeting summaries on topic: Abstractive multi-modal meeting
  summarization.
\newblock \emph{Proceedings of the 57th Annual Meeting of the Association for
  Computational Linguistics,}, pages 2190--2196.

\bibitem[{Lin(2004)}]{rouge}
Chin-Yew Lin. 2004.
\newblock Rouge: A package for automatic evaluation of summaries.
\newblock \emph{Text Summarization Branches Out,}.

\bibitem[{Lin et~al.(2015)Lin, Liu, Yang, Li, Zhou, and Li}]{docmodelhier}
Rui Lin, Shujie Liu, Muyun Yang, Mu~Li, Ming Zhou, and Sheng Li. 2015.
\newblock Hierarchical recurrent neural network for document modeling.
\newblock \emph{Proceedings of the 2015 Conference on Empirical Methods in
  Natural Language Processing,}, pages 899--907.

\bibitem[{Liu et~al.(2019{\natexlab{a}})Liu, Wang, Xu, Li, and Ye}]{didi}
Chunyi Liu, Peng Wang, Jiang Xu, Zang Li, and Jieping Ye. 2019{\natexlab{a}}.
\newblock Automatic dialogue summary generation for customer service.
\newblock In \emph{Proceedings of the 25th ACM SIGKDD International Conference
  on Knowledge Discovery \& Data Mining}, pages 1957--1965.

\bibitem[{Liu et~al.(2020)Liu, Jiang, He, Chen, Liu, Gao, and Han}]{Liu2020On}
Liyuan Liu, Haoming Jiang, Pengcheng He, Weizhu Chen, Xiaodong Liu, Jianfeng
  Gao, and Jiawei Han. 2020.
\newblock \href {https://openreview.net/forum?id=rkgz2aEKDr} {On the variance
  of the adaptive learning rate and beyond}.
\newblock In \emph{International Conference on Learning Representations}.

\bibitem[{Liu(2019)}]{bertsum}
Yang Liu. 2019.
\newblock Fine-tune bert for extractive summarization.
\newblock \emph{arXiv preprint arXiv:1903.10318}.

\bibitem[{Liu and Chen(2019)}]{liu-chen-2019-reading}
Zhengyuan Liu and Nancy Chen. 2019.
\newblock \href {https://doi.org/10.18653/v1/P19-1543} {Reading turn by turn:
  Hierarchical attention architecture for spoken dialogue comprehension}.
\newblock In \emph{Proceedings of the 57th Annual Meeting of the Association
  for Computational Linguistics}, pages 5460--5466, Florence, Italy.
  Association for Computational Linguistics.

\bibitem[{Liu et~al.(2019{\natexlab{b}})Liu, Ng, Lee, Aw, and
  Chen}]{liu2019topic}
Zhengyuan Liu, Angela Ng, Sheldon Lee, Ai~Ti Aw, and Nancy~F Chen.
  2019{\natexlab{b}}.
\newblock Topic-aware pointer-generator networks for summarizing spoken
  conversations.
\newblock \emph{arXiv preprint arXiv:1910.01335}.

\bibitem[{McCowan et~al.(2005)McCowan, Carletta, Kraaij, Ashby, Bourban, Flynn,
  Guillemot, Hain, Kadlec, Karaiskos et~al.}]{ami}
Iain McCowan, Jean Carletta, Wessel Kraaij, Simone Ashby, S~Bourban, M~Flynn,
  M~Guillemot, Thomas Hain, J~Kadlec, Vasilis Karaiskos, et~al. 2005.
\newblock The ami meeting corpus.
\newblock \emph{Proceedings of the 5th International Conference on Methods and
  Techniques in Behavioral Research,}, 88:100.

\bibitem[{Mehdad et~al.(2013)Mehdad, Carenini, Tompa et~al.}]{mehdad2013}
Yashar Mehdad, Giuseppe Carenini, Frank Tompa, et~al. 2013.
\newblock Abstractive meeting summarization with entailment and fusion.
\newblock \emph{Proceedings of the 14th European Workshop on Natural Language
  Generation,}, pages 136--146.

\bibitem[{Mihalcea and Tarau(2004)}]{textrank}
Rada Mihalcea and Paul Tarau. 2004.
\newblock Textrank: Bringing order into text.
\newblock \emph{Proceedings of the 2004 conference on empirical methods in
  natural language processing,}.

\bibitem[{Murray et~al.(2010)Murray, Carenini, and Ng}]{murray2010}
Gabriel Murray, Giuseppe Carenini, and Raymond Ng. 2010.
\newblock Generating and validating abstracts of meeting conversations: a user
  study.
\newblock \emph{Proceedings of the 6th International Natural Language
  Generation Conference,}, pages 105--113.

\bibitem[{Nallapati et~al.(2016)Nallapati, Zhou, dos Santos, Gu̇l{\c{c}}ehre,
  and Xiang}]{dailymail}
Ramesh Nallapati, Bowen Zhou, Cicero dos Santos, {\c{C}}a{\u{g}}lar
  Gu̇l{\c{c}}ehre, and Bing Xiang. 2016.
\newblock Abstractive text summarization using sequence-to-sequence rnns and
  beyond.
\newblock In \emph{Proceedings of The 20th SIGNLL Conference on Computational
  Natural Language Learning}, pages 280--290.

\bibitem[{Narayan et~al.(2018)Narayan, Cohen, and Lapata}]{xsum}
Shashi Narayan, Shay~B Cohen, and Mirella Lapata. 2018.
\newblock Don’t give me the details, just the summary! topic-aware
  convolutional neural networks for extreme summarization.
\newblock In \emph{Proceedings of the 2018 Conference on Empirical Methods in
  Natural Language Processing}, pages 1797--1807.

\bibitem[{Oya et~al.(2014)Oya, Mehdad, Carenini, and Ng}]{oya2014}
Tatsuro Oya, Yashar Mehdad, Giuseppe Carenini, and Raymond Ng. 2014.
\newblock A template-based abstractive meeting summarization: Leveraging
  summary and source text relationships.
\newblock \emph{Proceedings of the 8th International Natural Language
  Generation Conference (INLG),}, pages 45--53.

\bibitem[{Paulus et~al.(2018)Paulus, Xiong, and Socher}]{rlsummarize}
Romain Paulus, Caiming Xiong, and Richard Socher. 2018.
\newblock \href {https://openreview.net/forum?id=HkAClQgA-} {A deep reinforced
  model for abstractive summarization}.
\newblock In \emph{International Conference on Learning Representations}.

\bibitem[{Raffel et~al.(2019)Raffel, Shazeer, Roberts, Lee, Narang, Matena,
  Zhou, Li, and Liu}]{t5}
Colin Raffel, Noam Shazeer, Adam Roberts, Katherine Lee, Sharan Narang, Michael
  Matena, Yanqi Zhou, Wei Li, and Peter~J Liu. 2019.
\newblock Exploring the limits of transfer learning with a unified text-to-text
  transformer.
\newblock \emph{arXiv preprint arXiv:1910.10683}.

\bibitem[{Riedhammer et~al.(2008)Riedhammer, Gillick, Favre, and
  Hakkani-T{\"u}r}]{random}
Korbinian Riedhammer, Dan Gillick, Benoit Favre, and Dilek Hakkani-T{\"u}r.
  2008.
\newblock Packing the meeting summarization knapsack.
\newblock \emph{Ninth Annual Conference of the International Speech
  Communication Association,}.

\bibitem[{Rush et~al.(2015)Rush, Chopra, and Weston}]{nnlm}
Alexander~M Rush, Sumit Chopra, and Jason Weston. 2015.
\newblock A neural attention model for abstractive sentence summarization.
\newblock In \emph{Proceedings of the 2015 Conference on Empirical Methods in
  Natural Language Processing}, pages 379--389.

\bibitem[{Sandhaus(2008)}]{nyt}
Evan Sandhaus. 2008.
\newblock The new york times annotated corpus.
\newblock \emph{Linguistic Data Consortium, Philadelphia}, 6(12):e26752.

\bibitem[{See et~al.(2017)See, Liu, and Manning}]{pgnet}
Abigail See, Peter~J Liu, and Christopher~D Manning. 2017.
\newblock Get to the point: Summarization with pointer-generator networks.
\newblock In \emph{Proceedings of the 55th Annual Meeting of the Association
  for Computational Linguistics (Volume 1: Long Papers)}, pages 1073--1083.

\bibitem[{Serban et~al.(2016)Serban, Sordoni, Bengio, Courville, and
  Pineau}]{hred}
Iulian~V Serban, Alessandro Sordoni, Yoshua Bengio, Aaron Courville, and Joelle
  Pineau. 2016.
\newblock Building end-to-end dialogue systems using generative hierarchical
  neural network models.
\newblock \emph{Thirtieth AAAI Conference on Artificial Intelligence,}.

\bibitem[{Shang et~al.(2018)Shang, Ding, Zhang, Tixier, Meladianos,
  Vazirgiannis, and Lorr{\'e}}]{unsup18}
Guokan Shang, Wensi Ding, Zekun Zhang, Antoine Tixier, Polykarpos Meladianos,
  Michalis Vazirgiannis, and Jean-Pierre Lorr{\'e}. 2018.
\newblock Unsupervised abstractive meeting summarization with multi-sentence
  compression and budgeted submodular maximization.
\newblock In \emph{Proceedings of the 56th Annual Meeting of the Association
  for Computational Linguistics (Volume 1: Long Papers)}, pages 664--674.

\bibitem[{Sutskever et~al.(2014)Sutskever, Vinyals, and Le}]{seq2seq}
Ilya Sutskever, Oriol Vinyals, and Quoc~V Le. 2014.
\newblock Sequence to sequence learning with neural networks.
\newblock \emph{Advances in neural information processing systems,}, pages
  3104--3112.

\bibitem[{Vaswani et~al.(2017)Vaswani, Shazeer, Parmar, Uszkoreit, Jones,
  Gomez, Kaiser, and Polosukhin}]{transformer}
Ashish Vaswani, Noam Shazeer, Niki Parmar, Jakob Uszkoreit, Llion Jones,
  Aidan~N Gomez, {\L}ukasz Kaiser, and Illia Polosukhin. 2017.
\newblock Attention is all you need.
\newblock \emph{Advances in neural information processing systems,}, pages
  5998--6008.

\bibitem[{Wang and Cardie(2013)}]{wang2013}
Lu~Wang and Claire Cardie. 2013.
\newblock Domain-independent abstract generation for focused meeting
  summarization.
\newblock \emph{Proceedings of the 51st Annual Meeting of the Association for
  Computational Linguistics (Volume 1: Long Papers),}, 1:1395--1405.

\bibitem[{Zhang et~al.(2019)Zhang, Zhao, Saleh, and Liu}]{pegasus}
Jingqing Zhang, Yao Zhao, Mohammad Saleh, and Peter~J Liu. 2019.
\newblock Pegasus: Pre-training with extracted gap-sentences for abstractive
  summarization.
\newblock \emph{arXiv preprint arXiv:1912.08777}.

\bibitem[{Zhao et~al.(2019)Zhao, Pan, Fan, Liu, Li, Yang, and
  Cai}]{zhao2019abstractive}
Zhou Zhao, Haojie Pan, Changjie Fan, Yan Liu, Linlin Li, Min Yang, and Deng
  Cai. 2019.
\newblock Abstractive meeting summarization via hierarchical adaptive segmental
  network learning.
\newblock In \emph{The World Wide Web Conference}, pages 3455--3461.

\bibitem[{Zhu et~al.(2019)Zhu, Yang, Gmyr, Zeng, and Huang}]{leadbias}
Chenguang Zhu, Ziyi Yang, Robert Gmyr, Michael Zeng, and Xuedong Huang. 2019.
\newblock Make lead bias in your favor: A simple and effective method for news
  summarization.
\newblock \emph{arXiv preprint arXiv:1912.11602}.

\end{thebibliography}
\bibliographystyle{acl_natbib}
\clearpage
\appendix
\section{Training Details}
% The beam search width is 6 all our experiments. 

% We train the model for 50 epochs. The training is on a single Tesla V-100 GPU with 32G memory.
All the training is conducted on 4 Tesla V-100 GPU with 32G memory.
The batch size per GPU is 4096 tokens during pretraining and 8300 during finetuning.
The pretraining converges after $300,000$ steps, which runs for approximately $4$ days. The finetuning for both meeting datasets converges after $20,000$ steps, which runs for $6$ hours. We pick the model with the highest ROUGE-1 score on the development set of news and meeting datasets for pretraining and finetuning, respectively. 
%The ROUGE scores are computed using the open-source implementation \footnote{\urlhttp{https://github.com/andersjo/pyrouge/tree/master/tools/ROUGE-1.5.5}}.

% \begin{table*}[tpb]
% \resizebox{\textwidth}{!}{%
% \begin{tabular}{@{}c|ccc|ccc@{}}
% \toprule
%  & \multicolumn{3}{c|}{AMI} & \multicolumn{3}{c|}{ICSI} \\ \midrule
%  & R-1 & R-2 & R-SU4 & R-1 & R-2 & R-SU4 \\
% HMNet-1 & 49.30/51.74 & 16.80/18.59 & 23.06/24.23 & 47.55/44.84 & 9.63/10.75 & 18.87/18.40 \\
% HMNet-4-no-shuf & 50.11/52.93 & 17.20/18.62 & 23.32/24.80 & 45.99/44.99 & 9.90/10.68 & 17.94/18.42 \\
% HMNet-4 & 49.50/53.02 & 17.01/18.57 & 22.74/24.85 & 48.96/46.28 & 11.43/10.60 & 19.62/19.12 \\ \bottomrule
% \end{tabular}%
% }
% \caption{Development and test set performance for different pretraining variants on AMI and ICSI dataset.}
% \label{tab:pretrain_var}
% \end{table*}

% Due to the large computation cost of pretraining, we only run three variants of the pretraining. "HMNet-1" refers to $M=1$, where each pseudo meeting contains only a single news article. "HMNet-4" refers to our proposed pretraining scheme where $M=4$. "HMNet-4-no-shuf" is same as "HMNet-4" except that we do not shuffle the turns. The development and test set performance of these variants on AMI and ICSI is shown in table \ref{tab:pretrain_var}.

Due to the large computation cost of pretraining, we only tune hyperparameters for the decoding, namely the minimum length of the generated summary and beam size.
For both AMI and ICSI, we first set beam size to $3$ and grid search the minimum length from $\{240,280,320,360,400,440\}$. After selecting the best minimum length, we tune the beam size from $\{1,3,6,8,9,10\}$. The tuning is based on the development set ROUGE-1 score. 
The selected hyperparameters for AMI and ICSI and the corresponding development set performance is shown in table \ref{tab:best_hyper}.
All hyperparameter search trials and development/test set performance could be found in table \ref{tab:hyper-para-min-len} and \ref{tab:hyper-para-bsz}.

\begin{table}[]
\resizebox{0.5\textwidth}{!}{%
\begin{tabular}{c|ccccc}
\hline
 & Min. Len. & Beam Size & R-1 & R-2 & R-SU4 \\ \hline
AMI & 400 & 6 & 49.50 & 17.01 & 22.74 \\
ICSI & 280 & 6 & 48.96 & 11.34 & 19.62 \\ \hline
\end{tabular}%
}
\caption{Selected hyperparameters and development set rouge scores for the reported performance in table \ref{table:mainresult}}
\label{tab:best_hyper}
\end{table}

\begin{table}[]
\resizebox{0.5\textwidth}{!}{%
\begin{tabular}{c|cc|cc}
\hline
 & \multicolumn{2}{c|}{AMI} & \multicolumn{2}{c}{ICSI} \\ \hline
Min. Len. & R-1 (Dev) & R-1 (Test) & R-1 (Dev) & R-1 (Test) \\ \hline
240 & 47.41 & 52.82 & 45.98 & 45.68 \\
280 & 47.75 & 53.05 & \textbf{47.12} & 45.68 \\
320 & 48.03 & 53.11 & 46.88 & 45.68 \\
360 & 48.52 & 52.31 & 46.90 & 45.68 \\
400 & \textbf{48.68} & 51.40 & 46.27 & 46.10 \\
440 & 48.39 & 50.35 & 45.68 & 45.82 \\ \hline
\end{tabular}%
}
\caption{Hyperparameter search trials of minimum generation length with beam size fixed as 3. The bold numbers are the best development set performance with the selected minimum generation length.}
\label{tab:hyper-para-min-len}
\end{table}

\begin{table}[]
\resizebox{0.5\textwidth}{!}{%
\begin{tabular}{c|cc|cc}
\hline
 & \multicolumn{2}{c|}{AMI} & \multicolumn{2}{c}{ICSI} \\ \hline
Beam Size & R-1 (Dev) & R-1 (Test) & R-1 (Dev) & R-1 (Test) \\ \hline
1 & 48.09 & 50.97 & 45.13 & 45.54 \\
3 & 48.68 & 51.40 & 47.12 & 45.68 \\
6 & \textbf{49.50} & 53.02 & \textbf{48.96} & 46.28 \\
8 & 49.42 & N/A & 48.12 & 46.35 \\
9 & N/A & N/A & 48.35 & 45.45 \\
10 & N/A & N/A & 47.72 & 45.88 \\ \hline
\end{tabular}%
}
\caption{Hyperparameter search trials of beam size with minimum generation length fixed with 400 for AMI and 280 for ICSI. The bold numbers are the best development set performance with the selected beam size. "N/A" in the table is due to the GPU memory overflow issue for large beam size.}
\label{tab:hyper-para-bsz}
\end{table}

% Since \textbf{UNS} includes 4 model variants, we select the best result from all variants for comparison.

\section{Example of Meeting Summary} We demonstrate in Table~\ref{table:example1}, Table~\ref{table:example2} and Table~\ref{table:example3} examples of AMI meeting transcript with speaker information and three versions of summaries: reference, HMNet and UNS \cite{unsup18}. Since the transcript results are from ASR pipelines, there are some word errors and grammatical glitches. Moreover, compared with document summarization tasks like CNN/Daily Mail \cite{cnn,dailymail}, the meeting transcript is pretty long and lacks the important-information-first structure. All of these add to the complexity of meeting summarization tasks.

The summary generated by HMNet includes both individual actions/proposals and group activities, which is similar to the reference. In contrast, the result from UNS does not have a clear structure. Also, HMNet is more effective in selecting salient information from the lengthy transcript. Furthermore, the language of summary from HMNet is smoother and has many fewer grammatical errors than UNS. The reason is that HMNet learns the language pattern from the reference summary during training while UNS generates summaries by concatenating transcript word sequences.

% Example 1

\begin{table*}[htbp]
\centering
\small
\begin{tabular}{p{15cm}}
\midrule
\textbf{Meeting Transcript (163 turns)} \tabularnewline
\midrule
% TS3003b
ME: ... I've done some research. We have we have been doing research in a usability lab where we observed users operating remote controls. we let them fill out a questionnaire. Remotes are being considered ugly. F seventy five percent of the people questioned indicated that they thought their remote were was ugly. and an additional eighty percent indicated that they would spend more money on a fancy-looking remote control. Fifty percent of the people indicated they only loo used about ten percent of the buttons on a remote control ...\newline
ID: I've got a presentation about the working design. first about how it works. It's really simple. Everybody knows how a remote works. The user presses a button. The remote determines what button it is, uses the infrared to send a signal to the TV ... they only use about ten percent of the buttons, we should make very few buttons ...\newline
PM: ... another point is we have to skip the teletext, because in the world of upcoming internet we think teletext is going to be a thing of the past. And it's a function we don't need in our remote control...\newline
... \newline
UI: But Got many functions in one remote control, you can see, this is quite simple remote control. few buttons but This re remote control got a lot of buttons. people don't like it, so what I was thinking about was keep the general functions like they are.\newline
PM: Extra button info. that should be possible as. let's see what did we say. More. Should be fancy to, fancy design, easy to learn. Few buttons, we talked about that. Docking station, LCD. general functions And default materials... And we have to be very attent in putting the corporate image in our product. So it has to be visible in our design, in the way our device works... \newline
...\newline
PM: ... I will put the minutes in the project document folder... And we have a lunch-break now.  \tabularnewline
\midrule
\textbf{Reference Summary (14 sentences)} \tabularnewline
\midrule
The Project Manager stated the agenda and the marketing expert discussed what functions are most relevant on a remote, what the target demographic is, and what his vision for the appearance of the remote is. \newline
The Marketing Expert also brought up the idea to include a docking station to prevent the remote from getting lost and the idea to include an LCD screen.\newline
The User Interface Designer pushed for a user interface with large buttons, a display function, a touchscreen, and the capability of controlling different devices.\newline
The team then discussed teletext, the target demographic, the buttons the remote should have, the idea of marketing a remote designed for the elderly, an audio signal which can sound if the remote is lost, LCD screens, and language options.\newline
...\newline
Whether to include teletext in the design despite the new requirement which indicates that the team is not to work with teletext.\newline
The buttons are generally used, but the main feature is ugly and ugly. \newline
The remote will only have a few buttons. \newline
The remote will feature a small LCD screen. \newline
The remote will have a docking station.\newline
... \tabularnewline
\midrule
\textbf{Summary from HMNet (ours, 23 sentences)} \tabularnewline
\midrule
The User Interface Designer and the Industrial Designer presented the components of a remote control device.\newline
The Marketing Expert presented research on the working design and selling buttons for the next meeting. \newline
The Industrial Designer gave a presentation of the functions of the remote. \newline
The Project Manager announced that the project would not include a teletext feature. \newline
The Project Manager will post the minutes per cent of the minutes.\newline
The User Interface Designer will focus on the corporate image of the company.\newline
The group decided on features to include in the remote, to include an LCD screen, and a docking station to change the layout of the interface.\newline
...\newline
The remote will have buttons as few buttons as possible.\tabularnewline
Whether to include docking station selection for it. \newline
What functions the remote should be.\tabularnewline
\midrule
\textbf{Summary from UNS (8 sentences)} \tabularnewline
\midrule
Buttons we talked about the docking station LCD general functions\newline
Fancy design easy to learn few buttons on the right places \newline
Simple manner to put a lot of functions of the remote control \newline
Pricing we need a great deal of people would indicated that an LCD screen in the remote control would be preferred \newline
Focusing on elderly people or people forty plus they wanted to work \newline
Seventy five percent of the people indicated that the remote got lost in the room\newline
Minimum number of buttons the real buttons we have to use rebecca\newline
Required so most existing remote control simply\tabularnewline
\bottomrule
\end{tabular}
\caption{Example meeting transcript in AMI and summary from reference, HMNet, and UNS. Roles of participants are coded as follows: PM - project manager, ME - marketing expert, ID - industrial designer, UI - user interface designer. We manually capitalize some words in the summaries from HMNet and UNS for better demonstration.} %TS3003b
\label{table:example1}
\end{table*}

% Example 2

\begin{table*}[htbp]
\centering
\small
\begin{tabular}{p{15cm}}
\midrule
\textbf{Meeting Transcript (159 turns)} \tabularnewline
\midrule
% TS3003c
PM: ... It's the conceptual design meeting. And a few points of interest in this meeting are the conceptual specification of components. conceptual specification of design. And also trend-watching ...\newline
ME: Doh. I'm gonna inform you about the trend-watching I've done over the past few days. we've done some market research. We distributed some more enquetes, questionnaires ... \newline
UI: ... And we need some new a attractive functions which attract people for using it. it's like a speak speech recognition and a special button for selecting subtitles. and overall user-friendly. using large buttons ...\newline
ID: About the components design. for the energy source we can use a basic battery or, a as an optional thing, a kinetic energy, like in a watch ... for the casing, the manufacturing department can deliver a flat casing, single or double curved casing ... And the chip-set really should be advanced because ...\newline
... \newline
ID: Let's look at the flat case. It's from the side so it's rather normal. The the single curved so I'm not really what they're gonna look like, but it's something like this. this type should be better for you or better Should prevent repetitive strain injury a bit ... \newline
PM: I suggest the single curved, because maybe the curve is pretty good to put the screen in ... \newline
UI: And to put the buttons for changing the channel over here\newline
UI: Maybe it's possible to make this side like Let's see. Colour to make this side like the right colour... \newline
...\newline
PM: ... The user interface design, it's the same story. And product evaluation. So the Industrial Designer and User Interface Designer are going to work together on this one ...  \tabularnewline
\midrule
\textbf{Reference Summary (18 sentences)} \tabularnewline
\midrule
The Project Manager opened the meeting and recapped the decisions made in the previous meeting. \newline
The Marketing Expert discussed his personal preferences for the design of the remote and presented the results of trend-watching reports, which indicated that there is a need for products which are fancy, innovative, easy to use, in dark colors, in recognizable shapes, and in a familiar material like wood. \newline
The User Interface Designer discussed the option to include speech recognition and which functions to include on the remote.\newline
The Industrial Designer discussed which options he preferred for the remote in terms of energy sources, casing, case supplements, buttons, and chips.\newline
The team then discussed and made decisions regarding energy sources, speech recognition, LCD screens, chips, case materials and colors, case shape and orientation, and button orientation.\newline
...\newline
The Industrial Designer and User Interface Designer will work together. \newline
The remote will use a conventional battery and a docking station which recharges the battery. \newline
Whether to use kinetic energy or a conventional battery with a docking station which recharges the remote.
\tabularnewline
\midrule
\textbf{Summary from HMNet (ours, 21 sentences)} \tabularnewline
\midrule
The Project Manager opened the meeting by conceptual components and conceptual design.\newline
The Industrial Designer discussed the interior workings of a remote and suggested that the remote should feature speech recognition. \newline
The User Interface Designer presented two existing products and discussed the option to design a remote for the docking station. \newline
The Marketing Expert discussed research from trend watchers. \newline
The trend watchers have been consulted about energy sources, such as voice recognition, speech recognition, case recognition, and overall buttons.\newline
They also discussed the possibility of using a flat, double curved case, and double curved or double curved cases.  \newline
The group discussed the options for energy source, and energy sources. \newline
The designers will work together on the prototype evaluation.\newline
...\newline
Having an LCD screen on the remote which covers on the outside of the station.
Kinetic energy sources are needed to add extra buttons. \newline
If the case should be used, it would be fancy, they could make a simple case with plastic or single curved case.\tabularnewline
\midrule
\textbf{Summary from UNS (9 sentences)} \tabularnewline
\midrule
Older people like to shake your remote control the fresh
Changing channels button on the right side \newline
Time look easily get screen would held in making the remote control easier\newline
Leads us to some personal preferences the remote control\newline
People would pay more for speech recognition in a remote control\newline
Research about the designing a fly interface designer are going to work\newline
Trendwatchers i consulted advise that it should be the remote control on the docking station should be telephone shape so you could imagine\newline
Start by choosing a case\newline
Show it people like wood but it raised the price
\tabularnewline
\bottomrule
\end{tabular}
\caption{Example of meeting transcript and summary from reference, HMNet, and UNS. } %TS3003c
\label{table:example2}
\end{table*}

% Example 3

\begin{table*}[htbp]
\centering
\small
\begin{tabular}{p{15cm}}
\midrule
\textbf{Meeting Transcript (245 turns)} \tabularnewline
\midrule
% ES2004b
PM: ... I'll go over what we decided last meeting. \newline
ID:  this is the working design, presented by me. What the first thing question I asked was what are we trying to design?, a device which just sends the signal to the TV to change its state, whether that be the power, or the channel or the volume ... \newline
UI: ... But we are to make it unique so that people want to buy it, will this two features together. So what the concept is to have a flip-top model... Findings most people prefer us user-friendly rather than complex remote controls... I would make flip-top with a trendy design.\newline
ME: ... Eighty percent would spend more money when a remote control would look fancy... are prepared to spend more money for something that's a bit nicer looking... current remote controls do not match with the operating behaviour of the user overall. \newline
... \newline
ME: We asked those two questions, the table relates to both questions, so we didn't differentiate. Would you prefer an LCD screen, that's multi-function remote and would you pay more for speech recognition in a remote control? ... \newline
ME: Do we have to initially, looking at the findings here, focus on a younger age group initially and then broaden out the market later. Do we really have to go for everyone right away?  \newline
UI: We could focus on the biggest market. If say people between age group of twenty to thirty five are the biggest market?\newline
... \tabularnewline
\midrule
\textbf{Reference Summary (25 sentences)} \tabularnewline
\midrule
The Industrial Designer gave his presentation on the basic functions of the remote. \newline
He presented the basic components that remotes share and suggested that smaller batteries be considered in the product design. \newline
The User Interface Designer presented his ideas for making the remote easy-to-use; he discussed using a simple design and hiding complicated features from the main interface.\newline
The Marketing Expert presented the findings from a lab study on user requirements for a remote control device, and discussed users' demand for a simple interface and advanced technology.\newline
The Project Manager presented the new requirements that the remote not include a teletext function, that it be used only to control television, and that it include the company image in its design.\newline
he group narrowed down their target marketing group to the youth market.\newline
...\newline
The group then discussed the shell-like shape of the remote and including several different casing options to buyers. \newline
The remote will be bundled with a docking station to recharge the remote's batteries and a user-friendly instruction manual, and multiple casings will be made available. \newline
The limitations of the budget will restrict the development of some features; several of the features that the group wanted to include may have to be made simpler to decrease cost.
\tabularnewline
\midrule
\textbf{Summary from HMNet (ours, 25 sentences)} \tabularnewline
\midrule
The Project Manager recapped the decisions made in the previous meeting.\newline
The Industrial Designer discussed the interior workings of a remote and the team discussed options for batteries, volume control, and LCD screens. \newline
The Marketing Expert also found that users want a fancy look and feel, trendy, fashionable, and user design. \newline
The User Interface Designer presented findings from trend watching reports which indicated a need for products which are technologically innovative, are attractive to a user. \newline
The group decided to include a flip and an LCD screen, and discussed how well the interface would be incorporated into it.\newline
The team then discussed how to minimize the number of functions and what functions to include in their design. \newline
...\newline
They discussed making the remote a profit target group and whether to include an LCD, ease of noise, voice recognition, a locator function, and the LCD screen. \newline
The Marketing Expert will focus on making the product more user as possible to compensate younger consumers.\newline
The remote will only have a small, trendy design, and will have a few buttons for the next meeting.
\tabularnewline
\midrule
\textbf{Summary from UNS (9 sentences)} \tabularnewline
\midrule
Important that the project was accessible to wide range of consumers white age groups \newline
Remote you gotta press a button on top the tv and it beeps \newline
Seventy five percent of user find most remote controls \newline
Point about pressing the pound sign of the bleep are in the room \newline
Stick them in a program have to control with this remote control \newline
Hold in the palm of the hand set for all tv it here occur \newline
Fair amount i run that it last a long time \newline
Change the state of the tv all other appliances sending a signal \newline
Market research at
\tabularnewline
\bottomrule
\end{tabular}
\caption{Example of meeting transcript and summary from reference, HMNet, and UNS.} %ES2004b
\label{table:example3}
\end{table*}

\end{document}